\documentclass[10pt,letterpaper]{article}

\usepackage[utf8]{inputenc} 
\usepackage[T1]{fontenc}    
\usepackage{lmodern}        
\usepackage[english]{babel} 

\usepackage[round]{natbib} 

\makeatletter
\AtBeginDocument{%
	\@ifundefined{NAT@and}{}{%
		\renewcommand{\NAT@and}{\&}%
	}%
}
\makeatother

\usepackage{hyperref}
\usepackage{nameref}

\usepackage[top=0.85in, left=2.75in, footskip=0.75in, marginparwidth=2in]{geometry} 
\usepackage{lastpage,fancyhdr,graphicx}

\usepackage{amsmath}        
\usepackage{tabularx}       
\usepackage{booktabs}       
\usepackage{adjustbox}      

\usepackage{subcaption}     
\usepackage{wrapfig}        
\usepackage[pscoord]{eso-pic} 
\usepackage[fulladjust]{marginnote} 
\usepackage{sidecap}        
\usepackage{epstopdf}       
\usepackage{float}          

\usepackage{caption}        

\usepackage[right]{lineno}

\usepackage{indentfirst}    
\usepackage{enumitem}
\usepackage{titlesec}

\titleformat{\section}[block]
{\large\bfseries}
{\thesection}{1em}{}

\titleformat{\subsection}[block]
{\normalsize\bfseries}
{\thesubsection}{1em}{}

\fancyheadoffset[L]{2.25in}
\fancyfootoffset[L]{2.25in}
\reversemarginpar

\title{Peptidomic-Based Prediction Model for Coronary Heart Disease Using a Multilayer Perceptron Neural Network}
\author{Jesús Celis-Porras \\
	Instituto Tecnológico de Durango, Tecnológico Nacional de México \\
	\href{mailto:jcelis@itdurango.edu.mx}{\nolinkurl{jcelis@itdurango.edu.mx}}
}

\begin{document}
	\sloppy 
	
	\maketitle
	
	\section*{Abstract}
Coronary heart disease (CHD) is a leading cause of death worldwide and contributes significantly to annual healthcare expenditures. To develop a non-invasive diagnostic approach, we designed a model based on a multilayer perceptron (MLP) neural network, trained on 50 key urinary peptide biomarkers selected via genetic algorithms. Treatment and control groups, each comprising 345 individuals, were balanced using the Synthetic Minority Over-sampling Technique (SMOTE). The neural network was trained using a stratified validation strategy. Using a network with three hidden layers of 60 neurons each and an output layer of two neurons, the model achieved a precision, sensitivity, and specificity of 95.67\%, with an F1-score of 0.9565. The area under the ROC curve (AUC) reached 0.9748 for both classes, while the Matthews correlation coefficient (MCC) and Cohen’s kappa coefficient were 0.9134 and 0.9131, respectively, demonstrating its reliability in detecting CHD. These results indicate that the model provides a highly accurate and robust non-invasive diagnostic tool for coronary heart disease.

\textbf{\textbf{Keywords:}}
\begin{itemize}
	\item Coronary heart disease (CHD)
	\item Cardiology proteomics,
	\item Multilayer perceptron (MLP) neural network,
	\item Genetic algorithm feature selection,
	\item Non-invasive diagnostic,
	\item Machine learning in cardiology
\end{itemize}


\section{Introduction}
Non-invasive diagnosis of coronary artery disease (CAD) constitutes a key challenge for improving patient survival and quality of life. In this context, non-invasive methods for detecting coronary disease have become essential tools for the clinical evaluation of patients with suspected or known cardiovascular pathology, avoiding invasive procedures \citep{Vallee2019, Zreik2019b, Shorewala2021}. In the present study, we propose a model based on multilayer perceptron (MLP) neural networks as part of an automated system for diagnosing CAD, the most common form of cardiovascular disease. This model integrates proteomic biomarkers through artificial neural network algorithms \citep{Bom2019, Mondal2025}.

\subsection{Coronary artery disease diagnosis}

Coronary artery disease is caused by the progressive narrowing of the coronary arteries due to the accumulation of cholesterol and other lipid substances in their walls. When blood flow to the heart muscle is partially or completely blocked, angina or myocardial infarction may occur \citep{IQWiG2020}. Globally, CAD remains the leading cause of death, accounting for more than 9 million fatalities per year \citep{WHO2022}. The American Heart Association reports that in the United States alone, the prevalence of ischemic heart disease is approximately 18.2 million cases, with more than 370,000 annual deaths due to this cause \citep{AHA2025}.

\subsection{Methods for coronary artery disease diagnosis}

Non-invasive methods for detecting CAD—such as echocardiography, electrocardiography (ECG), coronary computed tomography angiography (CCTA), and nuclear imaging—are fundamental in diagnostic evaluation due to their safety, accessibility, and clinical utility \citep{Pontone2017, Edvardsen2022, Dhaladhuli2023, Kravarioti2025}. However, these tests have important limitations, including variability in study quality, bias in endpoint definition, and difficulties in integrating findings across multiple modalities \citep{Matta2021, Bytyci2025, Luz2005}. Moreover, their availability and effectiveness may be affected by factors such as cost, local expertise, patient claustrophobia, or the presence of incompatible medical devices, particularly in techniques like magnetic resonance imaging \citep{Edvardsen2022}. Test selection may also be influenced by pre-test probability and disease prevalence. In low-risk populations, diagnostic utility decreases, potentially increasing false positives and leading to unnecessary procedures or anxiety \citep{CleerlyHealth2025, Perna2020, Chin2012}. Finally, identifying patients without apparent risk factors who require intensive treatment remains a diagnostic challenge \citep{Raff2005}.
CAD may remain asymptomatic until a severe event occurs, such as acute myocardial infarction or cardiac arrest. Therefore, accurate detection methods are crucial for identifying risk levels and preventing heart disease. In this regard, models based on neural networks and other artificial intelligence techniques have been developed to enhance non-invasive diagnostic capability. For example, Cheung et al. (2023) developed a 3D deep learning classifier based on a ResNet-50 architecture to directly classify normal subjects and patients with coronary artery disease (CAD) from coronary CT angiography (CTCA) images, achieving a 21.43\% improvement in classification accuracy over state-of-the-art models. Their approach not only enhances diagnostic performance but also provides explainability aligned with clinicians' criteria, facilitating integration into clinical workflows. \citep{Cheung2023}. Similarly, Zreik et al. (2019a) used three-dimensional autoencoders and SVMs to analyze coronary computed tomography angiography (CCTA) images, achieving an AUC of approximately 0.87 \citep{Zreik2019a}. Additionally, Candemir et al. (2019) trained a 3D convolutional neural network to automatically detect coronary atherosclerosis in CCTA, with an accuracy near 90\% and an AUC of 0.91 \citep{Candemir2019}. More recently, Rehman et al. (2025) proposed a hybrid model based on artificial neural networks and particle swarm optimization (PSO), combined with balancing techniques such as SMOTE, which achieved 97\% accuracy using clinical data from the NHANES study \citep{Rehman2025}. Likewise, Yadav et al. (2022) developed a machine learning model to detect CAD using only non-invasive clinical parameters, such as vital signs, medical history, and lipid profiles, demonstrating practical utility in supporting medical decisions without invasive procedures \citep{Yadav2022}.

\subsection{Proteomics as a biomedical information source}	

For the development of AI-based models for disease diagnosis, precise descriptions of underlying biological processes are essential. In this context, proteomics, as a central discipline in biotechnology, provides a rich and detailed source of biomedical information \citep{Aebersold2018}.
Proteomics is the large-scale study of proteins, focusing on their structure and function. Proteins are vital for living organisms, serving as key components of cellular metabolic pathways \citep{Kellner1999}. Proteome analysis provides a dynamic picture of all proteins expressed under specific conditions at a given time. Systematic study and comparison of proteomes across different metabolic or pathological states allows the identification of proteins whose presence, absence, or alteration correlates with specific physiological stages. In pathological proteomics, this enables the identification of biomarkers for disease diagnosis or prognosis \citep{Wilkins1997}.
Proteomics is a relatively recent science that has grown thanks to the consolidation of mass spectrometry and the exponential increase in gene and protein data in databases \citep{Maccoss2005}. The combination of these techniques with peptide and protein fractionation and separation methods, such as two-dimensional polyacrylamide gel electrophoresis (2D-PAGE) and high-performance liquid chromatography (HPLC), has established proteomics as a tool for large-scale protein analysis since the mid-1990s \citep{Aebersold2003}.

\subsection{Proteomics and machine learning for diagnosis}		

The integration of proteomics with machine learning techniques, particularly neural networks, holds great potential for the future of non-invasive diagnosis, enabling early detection and personalized treatment strategies based on individual proteomic profiles \citep{Unterhuber2021, Smith2025}.
Neural networks have become a powerful tool for analyzing proteomic information, especially in diagnostic models using samples such as urine or plasma. These architectures, inspired by biological neural networks, can process large datasets and detect complex patterns that remain invisible to traditional statistical methods.
Several studies have demonstrated the effectiveness of multilayer perceptrons (MLPs) in predicting coronary disease using proteomic data. For example, high-risk coronary plaques have been predicted from selected plasma proteins \citep{Bom2019}, and neural network architectures have been optimized with ReLU activation functions and Adam algorithms, integrating oversampling techniques to improve diagnostic performance \citep{Mondal2025}. Additionally, models combining clinical data and hemodynamic indices have been developed to increase diagnostic sensitivity \citep{Vallee2019}, as well as ensemble learning strategies that outperform traditional single models \citep{Shorewala2021}. As research progresses, neural networks applied to proteomic data, particularly from accessible samples like urine, are expected to play an increasingly important role in early and non-invasive cardiovascular disease detection.

Therefore, the objective of this study is to develop a multilayer perceptron (MLP) neural network model trained with proteomic information from urine samples to support automated diagnosis of coronary artery disease.
\section{Materials and Methods}

\subsection{Materials}

The database was composed of peptides obtained from urine samples of patients from multiple sources: FLEMENGHO (Flemish Study on Environment, Genes, and Health Outcomes), Belgium, from 1985 to 2004; CACTI (Coronary Artery Calcification in Type I Diabetes), USA, from 2000 to 2002; AusDiab (Australian Diabetes, Obesity and Lifestyle Study), Australia, from 1999 to 2000; and Zwolle Outpatient Clinic, Netherlands, from 1998 to 2001 \citep{Wei2023}. The study included 82 cases (CAD) with a mean age of 65.6 ± 12.9 years and 345 controls with a mean age of 50.8 ± 15.9 years. Sample analysis was conducted using capillary electrophoresis coupled to high-resolution tandem mass spectrometry (CE-MS) \citep{Theodoridis2008,Kistler2009}, resulting in the identification of 5,605 peptides from the coronary artery disease proteome, which served as the reference database for this study.

\subsection{Balancing Case-Control Imbalance}

To ensure representativeness of the datasets and mitigate the imbalance between cases and controls, the Synthetic Minority Over-sampling Technique (SMOTE) was employed, increasing the 82 cases to 345 samples. This method generates new synthetic instances of the minority class through linear interpolations between each sample and its nearest neighbors, avoiding direct duplication of observations \citep{Chawla2002}. In this way, both the control group and the cases were augmented to achieve a sufficient size for balanced training, validation, and test sets. This strategy provided more representative partitions of the study population while enhancing the stability and generalization capacity of the deep multilayer perceptron (MLP) model used.

\subsection{Predictive Feature Selection} 

For optimal selection of predictive features, a wrapper-based approach was applied, integrating genetic algorithms (GA) \citep{Holland1975} with linear classifiers: Linear Discriminant Analysis (LDA) \citep{Fisher1936} and Linear Naive Bayes \citep{Lewis1998}.
The GA acted as a global optimization metaheuristic, efficiently exploring the search space of possible feature combinations. LDA and Linear Naive Bayes functioned as fitness evaluators, empirically assessing the predictive performance of each subset proposed by the GA. This collaboration between heuristic exploration and statistical validation allowed the identification of feature subsets that were both relevant and optimal for improving classifier accuracy. The features selected through this strategy were subsequently used to train the primary classifier.

\subsection{MLP Model Development Strategy} 

A multilayer perceptron (MLP) neural network model was implemented for supervised classification, with particular attention to training robustness in the presence of potentially imbalanced data. To mitigate bias toward majority classes, an oversampling technique was applied, replicating minority class instances until matching the number of samples in the majority class. This balancing step was essential to prevent systematic favoritism of the dominant class during learning.
Next, stratified 10-fold cross-validation was used, following the proposal by Kohavi \citep{Kohavi1995}, employing the \texttt{cvpartition} function with the \texttt{'Stratify'} option set to true. This strategy ensured that class proportions were maintained in both training and test sets of each fold, providing a more reliable assessment of model performance, especially in imbalanced contexts. Within each fold, the training set was further stratified into a training subset and a validation subset. This additional split allowed implementation of early stopping as proposed by Prechelt \citep{Prechelt1998}, configured to halt training if the validation error did not improve for six consecutive epochs (\texttt{max\_fail = 6}). This technique was crucial for preventing overfitting and improving the model's generalization capacity.
Regarding architecture, the MLP was configured with three hidden layers, each containing 60 neurons, and hyperbolic tangent (tansig) activation functions. The scaled conjugate gradient algorithm (trainscg) was used for training due to its computational efficiency with moderately sized datasets. A regularization rate of 0.1 was included to penalize excessive weights and reduce overfitting.

\subsection{Evaluation Metrics for the MLP} 

The performance of the multilayer perceptron (MLP) model was evaluated using multiple complementary approaches. Model performance was first assessed through confusion matrices and ROC curves, which provide insight into class-specific prediction accuracy and overall discriminative ability. Additionally, learning curves for both loss and accuracy were analyzed to monitor model convergence during training.
Quantitative evaluation was conducted using the following metrics: accuracy, sensitivity (recall), specificity, F1-score, Matthews correlation coefficient (MCC), and Cohen's Kappa. Accuracy measures the overall proportion of correct predictions, sensitivity evaluates the model’s ability to correctly identify positive cases, and specificity assesses correct identification of negative cases. The F1-score provides a balance between precision and recall, while MCC and Cohen's Kappa offer robust measures of agreement between predicted and true labels, accounting for class imbalance.
Together, these metrics provide a comprehensive assessment of the MLP’s predictive performance and reliability.

\newpage

\section{Results}
In developing the predictive model, high accuracy was achieved using the 50 most informative peptides, selected from a total of 5,605 entries in the database via a genetic algorithm with populations of 150 individuals over 250 generations. Multiple neural network architectures were evaluated to identify models with optimal predictive capacity for medical diagnosis. The highest performance was obtained with a multilayer perceptron (MLP) comprising three hidden layers of 60 neurons each, utilizing the tansig activation function. Model development adhered to the methodology described in the Methods section.
The results for one of the most robust models are presented below:

\begin{figure}[H]
	\centering
	\begin{subfigure}[b]{0.48\textwidth}
		\includegraphics[height=5cm]{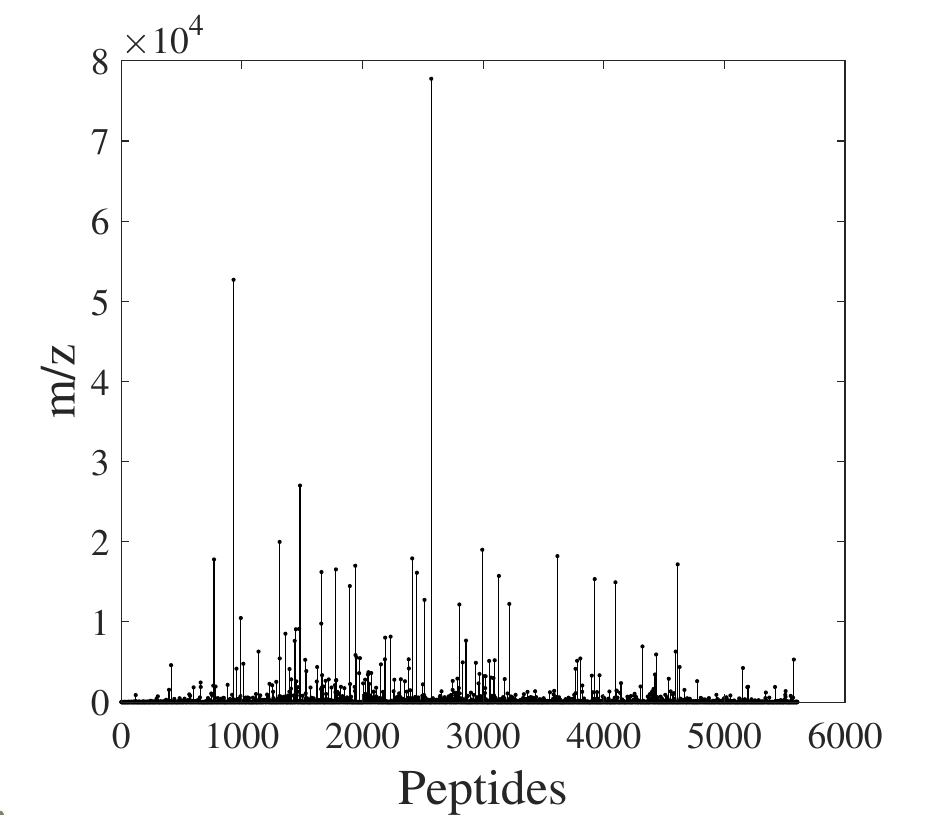}
		\caption{A case}
		\label{fig:case}
	\end{subfigure}
	\hfill
	\begin{subfigure}[b]{0.48\textwidth}
		\includegraphics[height=5cm]{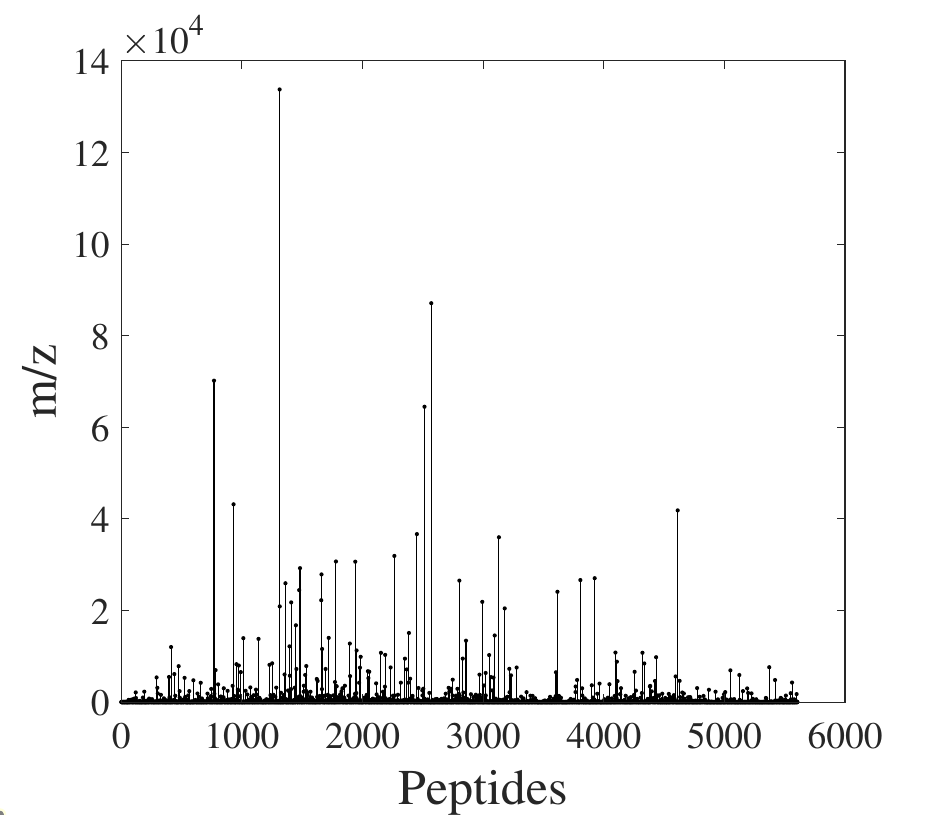}
		\caption{A control}
		\label{fig:control}
	\end{subfigure}
	\caption{Mass spectra of peptides for a case study (a) and a control (b).}
	\label{fig:case_control_main}
\end{figure}
Figure 1a: Mass spectrum of peptides (a case)   Figure 1b: Peptide mass spectrum (a control)
In Figure 1a (case sample), the spectrum is characterized by multiple peaks, with a maximum intensity of approximately 80,000 units ($8 \times 10^4$). Most signals are concentrated within the 1000–4000 peptide mass range, while only a few peaks exhibit higher intensity values. In Figure 1b (control sample), the spectrum shows a generally higher overall signal intensity. The most prominent peak exceeds 130,000 units ($13 \times 10^4$), accompanied by several other peaks of substantial magnitude distributed across the same peptide range. These spectral profiles serve as the basis for the subsequent comparative analysis between the case and control groups.

\begin{figure}[H]
	\centering
	\begin{subfigure}[b]{0.48\textwidth}
		\includegraphics[height=5cm]{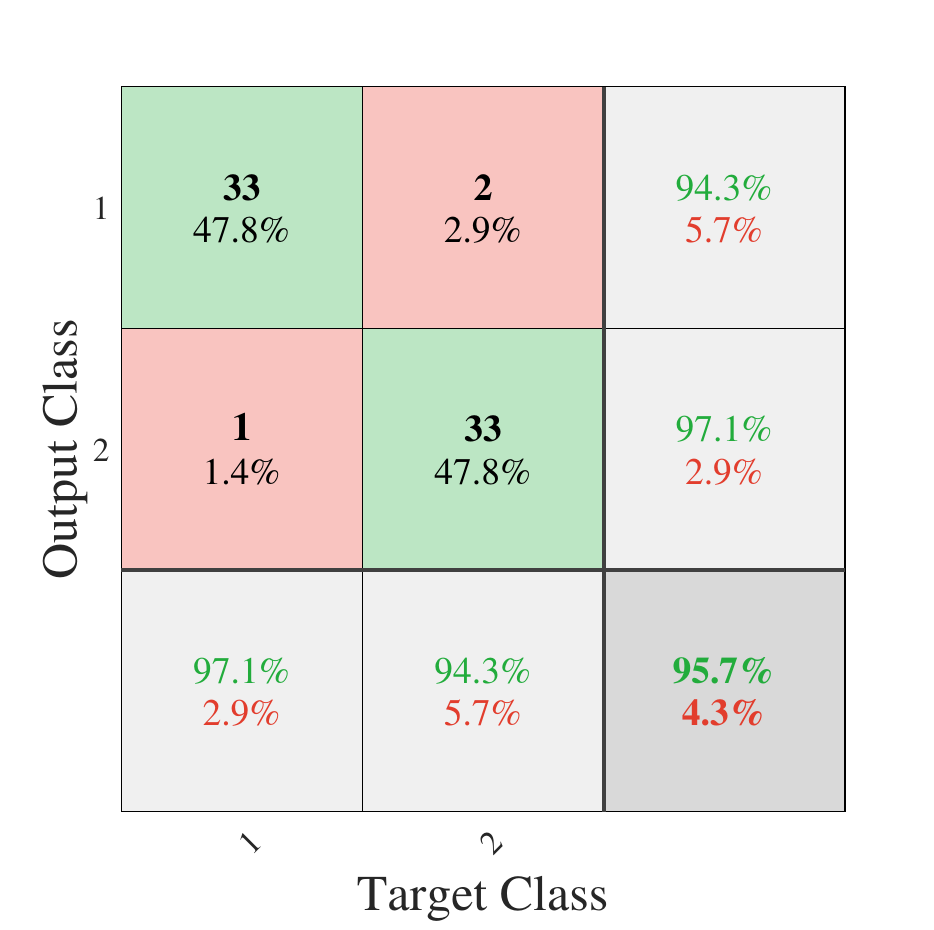}
		\caption{confusion matrix.}
		\label{fig:confusion_matrix}
	\end{subfigure}
	\hfill
	\begin{subfigure}[b]{0.48\textwidth}
		\includegraphics[height=5cm]{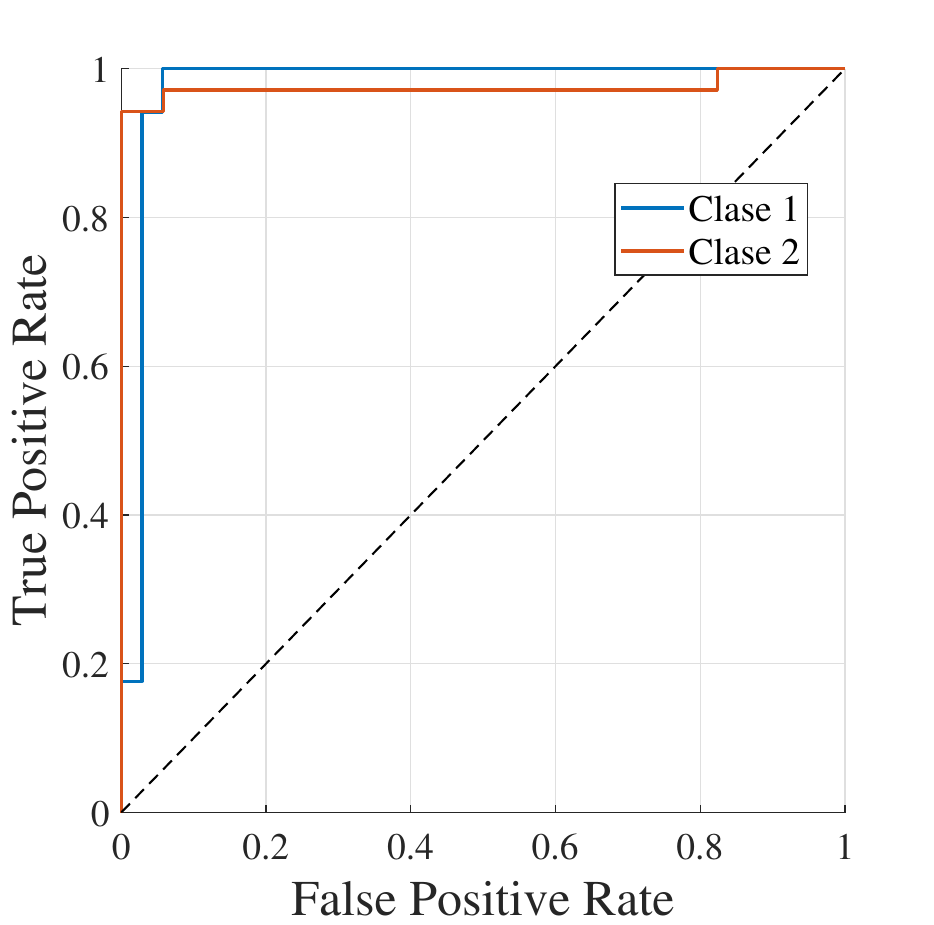}
		\caption{ROC curve.}
		\label{fig:roc_curve}
	\end{subfigure}
	\caption{Model Classification Results: Confusion Matrix (a) and ROC Curve (b).}
	\label{fig:confusion_roc_main}
\end{figure}

The confusion matrix (Figure 2a) shows that the model correctly classified 33 out of 34 instances in Class 1 (97.1\%) and 33 out of 35 instances in Class 2 (94.3\%), yielding an overall accuracy of 95.7\%. Misclassifications were limited to 2 instances of Class 1 and 1 instance of Class 2.
The Receiver Operating Characteristic (ROC) curve is presented in Figure 2b. The Area Under the Curve (AUC) for both Class 1 and Class 2 was 0.9748, Table 2 (see below). The macro-AUC was 0.9748, while the micro-AUC reached 0.9851, Table 3 (see below).

\begin{figure}[H]
	\centering
	\begin{subfigure}[b]{0.48\textwidth}
		\includegraphics[height=5cm]{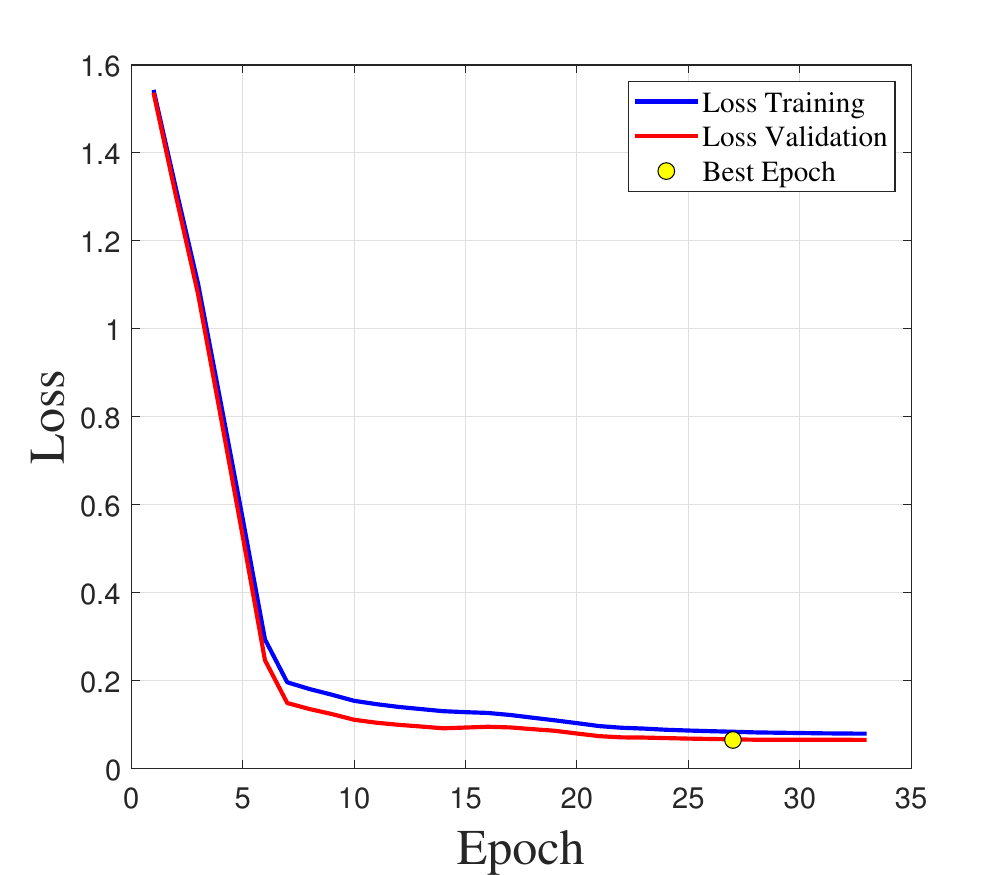}
		\caption{ Loss Curve Smoothed}
		\label{fig:Loss Curve Smoothed}
	\end{subfigure}
	\hfill
	\begin{subfigure}[b]{0.48\textwidth}
		\includegraphics[height=5cm]{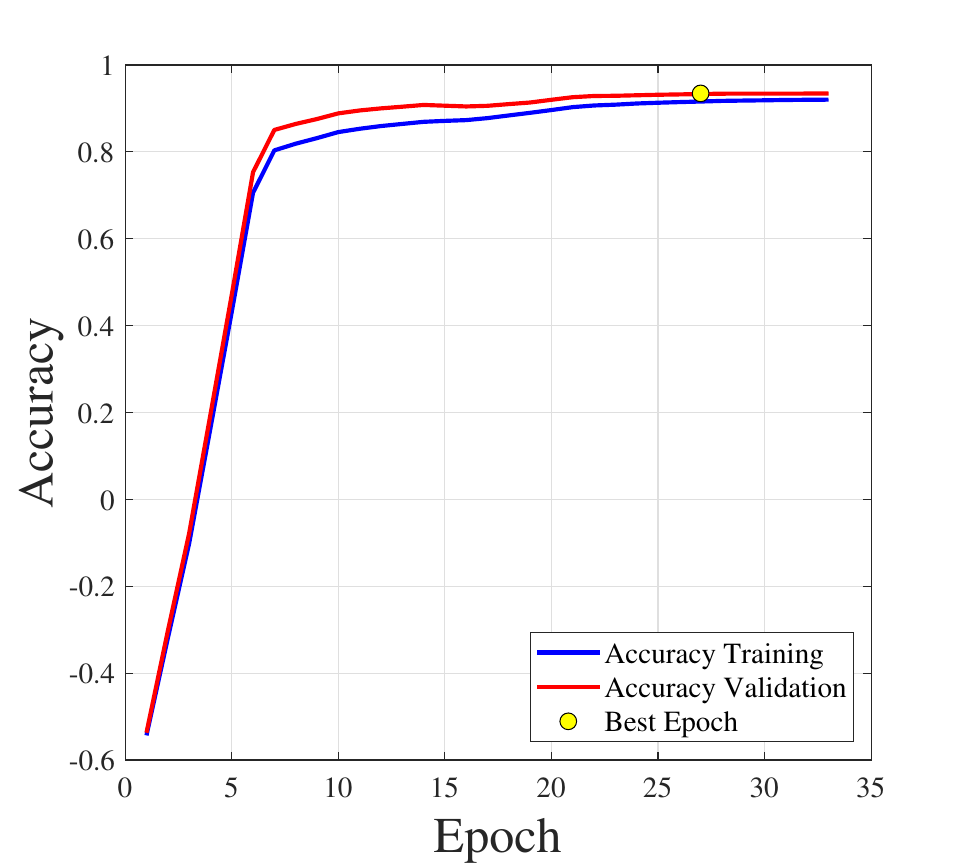}
		\caption{Accuracy Curve Smoothed}
		\label{fig:Accuracy Curve Smoothed}
	\end{subfigure}
	\caption{Model learning curves during the training phase.}
	\label{fig:learning-curves}
\end{figure}

Figure 3b shows the accuracy curves for training and validation. Accuracy increased rapidly during the first epochs, reaching approximately 80\% by epoch 5, and stabilized above 90\%, with the best validation performance observed at epoch 27.
Figure 3a illustrates the loss curves, which decreased from values above 1.5 at the beginning of training to approximately 0.1 at the final epoch. Training and validation loss followed a similar trend throughout.

\begin{table}[H]
	\centering
	\caption{Model performance metrics (CV and test set)}
	\label{tab:general_metrics}
	\begin{tabularx}{\textwidth}{l @{\extracolsep{\fill}} c}
		\toprule
		\textbf{Metric} & \textbf{Value} \\
		\midrule
		Mean accuracy (stratified cross-validation) & 90.72\% \\
		Final test accuracy & 95.65\% \\
		\bottomrule
	\end{tabularx}
\end{table}

Table 1 summarizes the classification performance of the multilayer perceptron (MLP) configured with three hidden layers of 60 neurons each. The average accuracy obtained through 10-fold stratified cross-validation was 90.72\%. The accuracy on the independent test set corresponding to the final fold was 95.65\%.

\begin{table}[H]
	\centering
	\caption{Model performance metrics on the test set.}
	\label{tab:class_metrics}
	\begin{tabularx}{\textwidth}{l c c}
		\toprule
		\textbf{Metric} & \textbf{Class 1 (Cases)} & \textbf{Class 2 (Controls)} \\
		\midrule
		Precision & 94.29\% & 97.06\% \\
		Sensitivity & 97.06\% & 94.26\% \\
		Specificity & 94.29\% & 97.06\% \\
		F1-Score & 95.65\% & 95.65\% \\
		AUC (Area Under the ROC Curve) & 0.9748 & 0.9748  \\
		\bottomrule
	\end{tabularx}
\end{table}
Table 2 summarizes the model’s performance in the current fold was evaluated using class-wise metrics, including precision, sensitivity (recall), specificity, and F1-score. For Class 1, the model achieved a precision of 0.9429, a sensitivity of 0.9706, a specificity of 0.9429, and an F1-score of 0.9565, indicating that the model correctly identifies the majority of cases in this class while maintaining a low rate of false positives.
For Class 2, the precision was 0.9706, the sensitivity 0.9429, the specificity 0.9706, and the F1-score 0.9565, demonstrating an equally balanced performance in classifying the controls.
These metrics indicate a consistent and balanced model performance across both classes, with high F1-scores suggesting a strong trade-off between precision and recall for each category.
\begin{table}[H]
	\centering
	\caption{Summary of model performance (averaged across classes)}
	\label{tab:average_metrics}
	\begin{tabularx}{\textwidth}{l @{\extracolsep{\fill}} c}
		\toprule
		\textbf{Metric} & \textbf{Average} \\
		\midrule
		Precision  & 95.67\% \\
		Sensitivity  & 95.67\% \\
		Specificity  & 95.67\% \\
		F1-Score & 95.65\%  \\
		Matthews Correlation Coefficient (MCC) & 0.9134  \\
		Cohen's Kappa & 0.9131 \\
		AUC Score macro (average per class) & 0.9748 \\
		AUC Score micro (weighted average) & 0.9851 \\
		\bottomrule
	\end{tabularx}
\end{table}
Table 3 summarizes the overall performance of the model, which was assessed using macro-averaged metrics across all classes, including precision, sensitivity (recall), specificity, F1-score, Matthews correlation coefficient (MCC), Cohen’s kappa coefficient, and area under the curve (AUC). The model achieved a macro-averaged precision, sensitivity, and specificity of 0.9567, along with an F1-score of 0.9565, indicating a balanced ability to correctly classify all classes.
Furthermore, the macro-averaged MCC was 0.9134, and the kappa coefficient was 0.9131, reflecting a high overall agreement between predicted and true labels beyond chance. Regarding discriminative ability, the macro-averaged AUC was 0.9748, while the micro-averaged AUC (weighted by class support) reached 0.9851, confirming strong model performance in distinguishing among classes.
These macro-averaged metrics demonstrate that the model maintains consistent and high performance across all classes, providing robust and reliable predictions.

\section{Discussion}
The combination of urinary proteomics, GA-based feature selection, and deep MLP networks provides a robust, non-invasive diagnostic approach for CHD. SMOTE effectively mitigated class imbalance, and stratified cross-validation with early stopping prevented overfitting. High values for AUC, MCC, and kappa indicate strong reliability and generalizability.
These findings support the potential clinical application of urinary peptide-based neural network models as complementary diagnostic tools for CHD, potentially reducing reliance on invasive procedures while maintaining high accuracy.

\subsection{Differential Peptidomic Profile}

Figures 1a and 1b show mass spectra representing peptide signal intensity (X-axis) versus mass-to-charge ratio (m/z, Y-axis). Both plots share the same peptide range (0–6000), allowing direct comparison. Notable differences are observed between them.
In Figure 1a, representing the proteome of a case group member, the spectrum shows a distribution of moderately intense peaks, with a dominant peak reaching approximately 80,000 units (8 $\times 10^4$). Most peaks are concentrated between 1000 and 4000 peptides, with intensity relatively low compared to the maximum peak.
Conversely, Figure 1b, representing a control group member, exhibits a more dispersed peak distribution and higher intensity signals. The highest peak exceeds 130,000 (13 $\times 10^4$), indicating higher relative abundance of that peptide. Several secondary peaks of considerable intensity suggest greater complexity or variability in the control sample.
Overall, the differences between Figures 1a and 1b indicate potential disparities in peptide expression between groups, which may reflect biologically relevant alterations between case and control subjects.

\subsection{Model Performance Metrics}
The multilayer perceptron (MLP) neural network, configured with three hidden layers of 60 neurons each, demonstrates outstanding performance in binary classification. The average accuracy of 90.72\% (see Table 1), obtained via 10-fold stratified cross-validation (k=10), robustly supports the model’s stability. Additionally, the 95.65\% accuracy (see Table 1) on the test set of the final fold confirms its strong generalization capability.

\subsection{Confusion Matrix-Based Evaluation of Model Performances}
The confusion matrix, Figure 2a, indicates that the model achieves a high and balanced performance across both classes, with precisions of 97.1\% for Class 1 and 94.3\% for Class 2 (see Table 2)
 yielding an overall precision of 95.67\% (see Table 3). Misclassifications were minimal, underscoring the model’s robustness, effective discriminative power, and generalizability. These results highlight its suitability for practical applications where reliable and unbiased classification is essential.

\subsection{ROC Curve Analysis}
Analysis of the Receiver Operating Characteristic (ROC) curve, Figure 2b, and the Area Under the Curve (AUC) demonstrates excellent discriminatory ability. An AUC of 0.9748 for both classes (see Table 2) indicates a high probability that the classifier ranks a positive instance above a negative one. The similarity between macro-AUC (0.9748) and micro-AUC (0.9851) (see Table 3) is consistent with the balanced class distribution after applying SMOTE (345 samples per class), confirming equitable performance across categories.

\subsection{Performance Metrics Analysis}

Class-specific metrics (see Table 2) from the final fold indicate, highly balanced performance:
\begin{itemize}
    \item Precision: Class 1 (0.9429) and Class 2 (0.9706) indicate high reliability in positive predictions.
    \item Recall: Class 1 (0.9706) and Class 2 (0.9429) demonstrate effectiveness in identifying positive cases.
    \item F1-Score: Both classes achieved 0.9565, reflecting an optimal balance between precision and recall.
\end{itemize}

\subsection{Classifier Robustness Evaluation}

Concordance coefficients, Matthews Correlation Coefficient (MCC), and Cohen's Kappa ($\kappa$) (see Table 3) provide a rigorous assessment of predictive quality:
\begin{itemize}
    \item MCC: 0.9134 indicating strong correlation between predictions and true labels.
    \item Kappa: 0.9131 demonstrating near-perfect agreement beyond chance.
\end{itemize}

\subsection{MLP Learning Analysis}

The stratified MLP exhibited progressive, stable learning with adequate generalization. Accuracy curves, Figure 3b, showed rapid improvement in early epochs, reaching ~80\% by epoch 5, and gradually stabilizing above 90\%, converging around epoch 27 (best validation point). The close alignment of training and validation curves indicates minimal overfitting.
Loss evolution, Figure 3a, corroborates this finding, decreasing sharply from initial values >1.5 to ~0.1 by the final phase. Balanced error reduction across training and validation confirms stable learning without compromising generalization.
Quantitative results reinforce these observations: average cross-validation accuracy was 90.72\%, while independent test performance reached 95.65\%. ROC curves reported high, balanced AUC for both classes (0.9748), with consistent metrics (F1-score = 0.9565, MCC = 0.9134, Kappa = 0.9131), evidencing a robust and reliable model.
Overall, loss and accuracy curve analysis concludes that the stratified MLP achieved efficient, stable learning, with consistent validation and test results, demonstrating effective handling of class imbalance via SMOTE and stratified cross-validation.
\subsection{MLP Model Analysis Based on Metric Values}
Results indicate that the MLP model is a robust and reliable classifier, based on consistent and high-quality performance metrics, (see Table 2). A classifier is considered robust when its performance does not significantly degrade under small variations in data or cross-validation folds. The low deviation between average accuracy (90.72\%) and test set accuracy (95.65\%) supports this claim. Robustness is further evidenced by stable class-specific metrics (precision, recall, F1-score).
Reliability is confirmed by Kappa (0.9131) and MCC (0.9134), which account for chance agreement. Values close to 1 indicate near-perfect alignment between predictions and true labels, ruling out random chance.
The MLP model can be described as exhibiting:
\begin{itemize}
    \item Remarkable generalization: The model performs well on unseen data, with test accuracy of 95.65\% confirming its ability to classify new samples effectively.
    \item Consistent performance across classes: Precision, recall, and F1-score are nearly identical for both classes (e.g., F1-score = 0.9565), indicating no class bias. ROC curves and AUC (0.9748) further confirm uniform discriminatory capability.
\end{itemize}
\section{Conclusions}
The main conclusions of the study, based on detailed model evaluation, are as follows:

\begin{enumerate}[label=\alph*)]
	\item Methodological robustness and reliability: Stratified cross-validation ensures unbiased performance estimation. Maintaining class proportions in each fold results in an average accuracy of 90.72\% (see Table 1), demonstrating strong generalization capability.
	\item High classification performance: The model exhibits exceptional classification ability, with key metrics including AUC = 0.9748 and average F1-score = 0.9565 (see Table 3), reflecting accurate distinction between classes. Precision, recall, and specificity, all near 95.7\%, further confirm reliability.
	\item Optimized model without apparent overfitting: The three-layer architecture with 60 neurons per layer, trainscg training, 0.1 regularization, and early stopping (6 epochs without improvement) prevented overfitting. High test metrics indicate the network learned underlying data patterns rather than memorizing the data.
	\item Cross-validation optimizes data usage: The training set was divided into training and validation subsets in each fold, allowing fine-tuning and optimal stopping without requiring a separate static validation set, maximizing model performance.
	\item Well-chosen network configuration: The selection of a network with three hidden layers of 60 neurons and the tansig activation function appears to be appropriate for the complexity of the problem. The network provides sufficient capacity to model the relationships within the data (as demonstrated by the high metrics) without being excessively complex. This results in a good balance between the model’s capacity and the risk of overfitting.
	\item Stable and efficient learning: The analysis of the loss and accuracy curves indicates that the stratified MLP achieved convergence in a stable manner, with consistent evolution of error and accuracy between training and validation. This behavior confirms that the model learned efficiently without overfitting, thereby reinforcing the reliability of the reported metrics and the model’s generalization capability.
\end{enumerate}

\bibliographystyle{apalike} 
\bibliography{CAD2}
			
\end{document}